\documentclass[10pt, letterpaper, journal, twoside]{IEEEtran}

\IEEEoverridecommandlockouts    
   
\usepackage{graphics} 
\usepackage{cite}
\usepackage[T1]{fontenc}
\usepackage[utf8]{inputenc}
\usepackage{amsmath, bm} 
\usepackage{amssymb}  
\usepackage{hyperref}
\hypersetup{
    colorlinks=true,
    linkcolor=blue,
    filecolor=magenta,      
    urlcolor=blue,
}

\usepackage{float}
\usepackage{cancel}

\usepackage{numprint}
\usepackage{comment}

\usepackage{pgfplots}
\usepackage{amssymb}
\pgfplotsset{compat=newest}
\usetikzlibrary{plotmarks}
\usetikzlibrary{arrows.meta}
\usepgfplotslibrary{patchplots}
\usepackage{amsmath}

\usepackage{color, colortbl}
\definecolor{LightCyan}{rgb}{0.88,1,1}
\usepackage{multirow}
\usepackage{siunitx}
\DeclareSIUnit\cell{cell}
\DeclareSIUnit\cells{cells}
\DeclareSIUnit\trees{trees}

\usepackage{titlesec} 
\renewcommand{\thesection}{\Roman{section}}
\renewcommand{\thesubsection}{\thesection-\Alph{subsection}}
\renewcommand{\thesubsubsection}{\thesubsection\arabic{subsubsection}}
\titleformat{\subsubsection}[runin]{\itshape}{\arabic{subsubsection})}{0.5em}{}

\titlespacing*{\subsubsection}{\parindent}{0pt}{*1}
\titlespacing*{\section}{0pt}{*1}{*1}
\titlespacing{\subsection}{0pt}{*1}{*1}

\usepackage{enumitem} 

\usepackage{wrapfig}
\usepackage{amssymb}
\usepackage{tikz}
\usetikzlibrary{shapes,arrows}
\usepackage{verbatim}
\usetikzlibrary{positioning}
\usepackage[normalem]{ulem} 
\usepackage{tabstackengine}
\usepackage{algorithm}
\usepackage[end]{algpseudocode} 
\usepackage{etoolbox}
\usepackage[font=small, skip=5pt]{caption}
\usepackage[skip=3pt]{subcaption} 

\usepackage[nodisplayskipstretch]{setspace}
\usepackage{pgfplots}
\usepackage{grffile}
\usetikzlibrary{plotmarks}
\usetikzlibrary{arrows.meta}
\usepgfplotslibrary{patchplots}
\usepackage{amsmath}

\usepackage{bbding} 

\usepackage{colortbl}

\usepackage{hhline}
\usepackage{makecell}

\usepackage{eso-pic}
\newcommand\AtPageUpperMyright[1]{\AtPageUpperLeft{%
 \put(\LenToUnit{0.5\paperwidth},\LenToUnit{-1.5cm}){%
     \parbox{0.5\textwidth}{\raggedright\fontsize{11}{11}\selectfont #1}}%
 }}%
\newcommand{\conf}[1]{%
\AddToShipoutPictureBG*{%
\AtPageUpperMyright{#1}
}
}

\title{\LARGE \bf 
GP-guided MPPI for Efficient Navigation in Complex Unknown Cluttered Environments
}

\conf{\hspace*{-9cm}\textcolor{gray}{This paper has been accepted for publication at the IEEE/RSJ International Conference on Intelligent Robots and Systems \textcolor{blue}{\href{https://ieee-iros.org/}{(IROS)}}, Detroit, Michigan, USA, 2023. }}

\author{Ihab S. Mohamed$^{*}$, Mahmoud Ali$^{*}$, and Lantao Liu
\thanks{Authors are with the Luddy School of Informatics, Computing, and Engineering, Indiana University, Bloomington, IN 47408 USA (e-mail: {\tt\small \{mohamedi, alimaa, lantao\}@iu.edu})\\
$^{*}$Ihab S. Mohamed and Mahmoud Ali
 equally contributed to this work. \\
 This work is supported by National Science Foundation with grant numbers 2006886 and 2047169. 
 }
}%
\definecolor{applegreen}{rgb}{0.8, 1, 0.0}
\definecolor{LightCyan}{rgb}{0.88,1,1}
\definecolor{atomictangerine}{rgb}{1.0, 0.6, 0.4}
\definecolor{amber}{rgb}{1.0, 0.75, 0.0}
\definecolor{aqua}{rgb}{0.0, 1.0, 1.0}
\definecolor{almond}{rgb}{0.94, 0.87, 0.8}
\definecolor{aquamarine}{rgb}{0.5, 1.0, 0.83}
\definecolor{babyblue}{rgb}{0.54, 0.81, 0.94}
\definecolor{babyblueeyes}{rgb}{0.63, 0.79, 0.95}
\definecolor{asparagus}{rgb}{0.53, 0.66, 0.42}
\definecolor{auburn}{rgb}{0.43, 0.21, 0.1}
\definecolor{brilliantlavender}{rgb}{0.96, 0.73, 1.0}
\definecolor{bittersweet}{rgb}{1.0, 0.44, 0.37}
\definecolor{blue-violet}{rgb}{0.54, 0.17, 0.89}
\definecolor{capri}{rgb}{0.0, 0.75, 1.0}
\definecolor{celadon}{rgb}{0.67, 0.88, 0.69}
\definecolor{darkcyan}{rgb}{0.0, 0.55, 0.55}
\definecolor{deepskyblue}{rgb}{0.0, 0.75, 1.0}
\definecolor{dogwoodrose}{rgb}{0.84, 0.09, 0.41}

\begin{document}

\maketitle

\global\csname @topnum\endcsname 0
\global\csname @botnum\endcsname 0


\thispagestyle{empty}
\pagestyle{empty}


\begin{abstract}
Robotic navigation in unknown, cluttered environments with limited sensing capabilities poses significant challenges in robotics. Local trajectory optimization methods, such as Model Predictive Path Intergal (MPPI), are a promising solution to this challenge. However, global guidance is required to ensure effective navigation, especially when encountering challenging environmental conditions or navigating beyond the planning horizon.
This study presents the GP-MPPI, an \textit{online} learning-based control strategy that integrates MPPI with a local perception model based on Sparse Gaussian Process (SGP). 
The key idea is to leverage the learning capability of SGP to construct a variance (uncertainty) surface, which enables the robot to learn about the navigable space surrounding it, identify a set of suggested subgoals, and ultimately recommend the optimal subgoal that minimizes a predefined cost function to the local MPPI planner. 
Afterward, MPPI computes the optimal control sequence that satisfies
the robot and collision avoidance constraints.
Such an approach eliminates the necessity of a global map of the environment or an offline training process.
We validate the efficiency and robustness of our proposed control strategy through both simulated and real-world experiments of 2D autonomous navigation tasks in complex unknown environments, demonstrating its superiority in guiding the robot safely towards its desired goal while avoiding obstacles and escaping entrapment in local minima. The GPU implementation of GP-MPPI, including the supplementary video, is available at \url{https://github.com/IhabMohamed/GP-MPPI}.
\end{abstract}
\vspace*{-3pt}
\vspace*{-1pt}
\section{Introduction and Related Work}\label{Introduction}
Autonomous navigation of mobile robots 
in unknown, cluttered, and unpredictable environments with limited sensor capabilities is a challenging task owing to the inherent uncertainty and complexity of such environments.
To tackle this challenge, a \textit{receding-horizon} strategy such as Model Predictive Control (MPC) is commonly employed. 
The MPC control framework allows the robot to simultaneously plan a short trajectory (sequence of actions), following which the robot executes the immediate action while planning a subsequent trajectory.
To successfully achieve receding-horizon planning, the robot must consider both safety and persistent feasibility, where \textit{safety} is achieved by avoiding collisions with any obstacles while executing a planned trajectory, and \textit{persistent feasibility} is maintained by always 
generating a safe trajectory that does not result in dead-ends or local minima while progressing towards the desired goal.
\begin{figure}%
    \centering
   \resizebox{1.03\columnwidth}{!}
    {
        \includegraphics[scale=0.7]{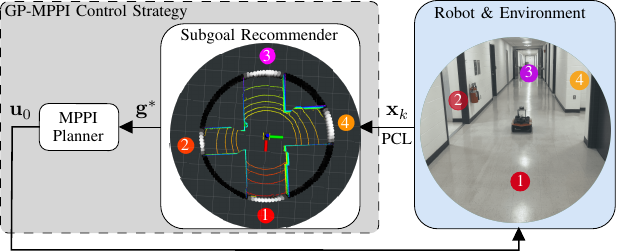}
    }
    \caption{Architecture of our proposed GP-MPPI control strategy, which comprises two main components: the GP-subgoal recommender and the local planner, the MPPI. First, the GP-subgoal recommender observes the surrounding environment and suggests the optimal subgoal position $\mathbf{g}^\ast$ to the local motion planner, where four colored circles represent the GP-recommended subgoals. MPPI then computes the optimal control sequence, which minimizes the distance to $\mathbf{g}^\ast$ while avoiding collision with obstacles and respecting system constraints, followed by executing the first optimal control $\mathbf{u}_{0}$ to the robot. 
    }%
    \label{fig:indoor-cluttered-enviroment}%
\end{figure}

One of the significant challenges in robot motion planning is that the desired goal is often situated beyond the planning horizon, which requires the use of local subgoals or \textit{cost-to-go} heuristics 
for motion safety and persistent feasibility. 
A common strategy is to rely on single-query motion planning algorithms, such as $\text{A}^{\!*}$ and $\text{RRT}^\text{X}$, to identify feasible paths that direct the local planner towards its desired goal \cite{koenig2005fast, otte2016rrtx}.
For instance, the $\text{RRT}^\text{X}$ algorithm, introduced in \cite{otte2016rrtx}, incorporates replanning techniques from Dynamic
Rapidly-exploring Random Trees (DRRT) and Rapid-exploring Random Trees ($\text{RRT}^*$) algorithms to adjust the path during exploration based on environmental changes. However, due to its high computational demands, implementing this algorithm in \textit{real-time} on a robot can be challenging.

One alternative method to achieve efficient solutions for motion planning problems 
is the integration of MPC with data-driven methods, also known as learning-based MPC \cite{hewing2020learning}.
To name a few, a subgoal planning policy using Deep Reinforcement Learning (DRL) is recently proposed to guide the local MPC planner to navigate in crowded surroundings \cite{brito2021go, lodel2022look}. 
Similarly, RL was utilized to choose the next subgoal from a set of predefined possibilities \cite{greatwood2019reinforcement}, which guides the robot through challenging environments with dead-end corridors while also prevents the MPC planner from getting trapped in local minima.
Another related work that combines learning with MPC is POLO which aims to enhance MPC performance by learning a global value function~\cite{lowreyplan}.
Most of these approaches typically rely on either offline training or having access to the global map of the environment.
In addition, many recent studies have suggested combining Gaussian Process (GP) with MPC to learn system dynamics, leading to better control performance and robustness to uncertainty \cite{wang2023learning}.

Another research avenue employed gap-based techniques that identify gaps as \textit{free} spaces between obstacles, enabling a robot to move through them while avoiding local minima and obstacles.
The first developed method was the Nearness Diagram (ND)~\cite{1266644}, but many of its variants exhibited undesired oscillatory motion. 
To overcome these limitations, robotics researchers have developed techniques that rely on the geometry of the gap. One such technique is the Follow-the-Gap Method (FGM), which selects a gap based on its area and computes the robot's heading using the gap center's direction relative to both the robot and the final goal \cite{sezer2012novel}.
Another approach is the sub-goal seeking method, which assigns a cost to each sub-goal based on the goal heading error with respect to the robot and the gap heading, and then selects the sub-goal with the lowest cost (error) \cite{ye2009sub}.
The Admissible Gap (AG) method \cite{mujahed2018admissible}, an iterative algorithm that takes into account the exact shape and kinematic constraints of the robot, identifies possible admissible gaps, and selects the nearest gap as the goal.
\raggedbottom 

Different from all these strategies, our proposed framework leverages a Sparse variant of Gaussian Process (SGP) which is a new perception model by ``abstracting'' local perception data so that the local sub-goal for navigation can be naturally extracted. 
Specifically, we introduce the GP-MPPI control strategy, which enhances the state-of-the-art sampling-based MPC, Model Predictive Path Integral (MPPI) \cite{williams2017model}, by incorporating the GP-subgoal recommender policy. Such a policy takes advantage of the SGP occupancy model to learn about the navigable space surrounding the robot, identifies a set of suggested subgoals, and ultimately recommends the optimal subgoal that minimizes a predefined cost function to the MPPI local planner, as demonstrated in Fig.~\ref{fig:indoor-cluttered-enviroment}.
Subsequently, MPPI computes the optimal control sequence that satisfies the robot and collision avoidance constraints while moving towards the recommended subgoal, followed by executing the first optimal control $\mathbf{u}_0$ to the robot.
In summary, the contributions of this work can be summarized as follows:
\begin{enumerate}
    \item We propose an \textit{online} learning-based control strategy that recommends subgoals solely based on local sensory information, ensuring safety and persistent feasibility; such an approach eliminates the need for a global map of the environment or an offline training process as in RL techniques, resulting in a more flexible and agile control framework that can be easily deployed in different unexplored environments, as revealed in Section~\ref{GP-MPPI Control Strategy}.
    \item To the best of the authors' knowledge, this is the first attempt to utilize the SGP occupancy model in conjunction with sampling-based trajectory optimization methods, specifically MPPI, to efficiently explore the navigable space surrounding the robot.
    \item In Sections~\ref{Simulation Details and Results} and \ref{Real-World Demonstration}, we validate our GP-MPPI control strategy for collision-free navigation in complex and unknown cluttered environments, using both simulation and experimental demonstrations; by comparing it with two baseline sampling-based approaches (namely, MPPI \cite{williams2017model}, and log-MPPI \cite{mohamed2022autonomous}), we show its effectiveness in overcoming local minima that may arise when the sampled trajectories of MPPI are concentrated in high-cost regions or due to challenging environmental conditions.
\end{enumerate}
\section{Preliminaries}\label{Preliminaries}
To provide the necessary background for our proposed work, in this section, we formulate the optimal control problem and present a concise overview of the MPPI control strategy that can be utilized to address this problem, along with a brief introduction to the Sparse Gaussian Process (SGP) which is the backbone of our GP-subgoal recommender policy. 
\subsection{Problem Formulation}
Consider a nonlinear discrete-time stochastic dynamical system
\vspace{-5pt}
\begin{equation}\label{dynamics_system}
   \mathbf{x}_{k+1}=f\left(\mathbf{x}_{k},\mathbf{u}_{k}+\delta \mathbf{u}_{k}\right),   
\end{equation}
with $\mathbf{x}_k \in \mathbb{R}^{n_x}$ and $\mathbf{u}_k \in \mathbb{R}^{n_u}$ representing the state of the system and its control input, respectively. The disturbance introduced into the control input, $\delta \mathbf{u}_{k}$, is modeled as a zero-mean Gaussian noise with co-variance $\Sigma_{\mathbf{u}}$.
Given a finite time-horizon $N$, we define the control sequence $\mathbf{U}$ as $\mathbf{U} = \left[\mathbf{u}_{0}, \mathbf{u}_{1}, \dots,\mathbf{u}_{N-1}\right]^{\top} \in \mathbb{R}^{n_u N}$ and the resulting state trajectory of the system being controlled as  $\mathbf{X} = \left[\mathbf{x}_{0}, \mathbf{x}_{1}, \dots, \mathbf{x}_{N}\right]^{\top} \in \mathbb{R}^{n_x (N+1)}$. 
Furthermore, $\mathcal{X}^d$ is used to represent the $d$-dimensional space with $\mathcal{X}_{rob}\left(\mathbf{x}_{k}\right) \subset \mathcal{X}^d$ and $\mathcal{X}_{o b s}\subset \mathcal{X}^d$ representing the robot's occupied area and obstacles' area, respectively. 
Let $\mathbf{x}_s$ and $\mathbf{x}_f$ denote the initial and desired (goal) state of the robot, respectively.
Given $\mathcal{X}_{rob}\left(\mathbf{x}_{k}\right), \mathcal{X}_{o b s}, \mathbf{x}_s$, and $\mathbf{x}_f$, we aim to find the optimal control sequence, $\mathbf{U}$, that allows the robot to safely and efficiently navigate from its initial state, $\mathbf{x}_s$, to the desired state, $\mathbf{x}_f$, by avoiding both getting stuck in local minima and collisions with obstacles, while minimizing a cost function $J$.
The optimization problem at hand can be approached utilizing the classical MPPI control strategy described in \cite{williams2017model}.  
This optimization can be mathematically expressed as in \eqref{eq:2}, with the objective of minimizing the cost function, $J$, which is comprised of the expectation of a combination of state terminal cost $\phi(\mathbf{x}_{N})$, running cost  $q(\mathbf{x}_{k})$, and control inputs $\mathbf{u}_{k}$, weighted by the positive-definite matrix $R\in \mathbb{R}^{n_u \times n_u}$, taking into consideration the system dynamics outlined in (\ref{eq:2b}) and constraints such as collision avoidance and control constraints as stated in (\ref{eq:2c}). 
\begin{subequations}
\begin{align}
\min _{\mathbf{U}} \quad J &=  \mathbb{E}\left[\phi\left(\mathbf{x}_{N}\right)+\sum_{k=0}^{N-1}\left(q\left(\mathbf{x}_{k}\right)+\frac{1}{2} \mathbf{u}_{k}^\top R \mathbf{u}_{k}\right)\right]\!, \label{eq:2a}\\
\text {s.t.} 
\quad & \mathbf{x}_{k+1}=f\left(\mathbf{x}_{k}, \mathbf{u}_{k}+\delta \mathbf{u}_{k}\right), \delta \mathbf{u}_{k} \sim \mathcal{N}(\mathbf{0}, \Sigma_{\mathbf{u}}), \label{eq:2b}\\
& \mathcal{X}_{rob}\left(\mathbf{x}_{k}\right) \cap \mathcal{X}_{obs}=\emptyset, \;\mathbf{h}(\mathbf{x}_k, \mathbf{u}_k) \leq 0, \label{eq:2c}\\
& \mathbf{x}_0 = \mathbf{x}_s, \;\mathbf{u}_{k} \in \mathbb{U},\; \mathbf{x}_{k} \in \mathbb{X}. 
\end{align}
\label{eq:2}
\vspace*{-10pt}
\end{subequations}
\subsection{Overview of MPPI Control Strategy}\label{Overview of MPPI Control Strategy}
In order to solve the optimization control problem defined in (\ref{eq:2}), MPPI leverages Monte Carlo simulation to generate a significant number of \textit{real-time} simulated trajectories by propagating them from the underlying system dynamics. It then evaluates the \textit{cost-to-go} of each trajectory based on a predefined cost function and updates the optimal control sequence by considering a weighted average cost from all of the simulated trajectories. More details are given in \cite{williams2017model, mohamed2022autonomous}.
Subsequently, each trajectory $\tau_i$ in the time-horizon $N$ can have its \textit{cost-to-go} evaluated as given in (\ref{eq:cost-to-go}), where the \textit{cost-to-go} $\tilde{S}(\tau_i)$ is calculated as the sum of the terminal state cost $\phi(\mathbf{x}_N)$ and the instantaneous running cost $\tilde{q}(\mathbf{x}_{k}, \mathbf{u}_{k}, \delta \mathbf{u}_{k,i})$ over all time steps. 
The instantaneous running cost, $\tilde{q}$, expressed in (\ref{eq: state-dependent cost-to-go}), is comprised of the state-dependent running cost $q(\mathbf{x}_{k})$ and the quadratic control cost $q(\mathbf{u}_{k}, \delta \mathbf{u}_{k})$, where $\gamma_\mathbf{u} = \frac{\nu -1}{2\nu}$ and the aggressiveness in exploring the state-space is determined by the parameter $\nu \in \mathbb{R}^{+}$. Specifically, 
\begin{equation}
\label{eq:cost-to-go}
 \tilde{S}\left(\tau_i \right) =\phi\left(\mathbf{x}_N\! \right) + \! \! \sum_{k=0}^{N-1} \!\tilde{q}\left(\mathbf{x}_{k}, \mathbf{u}_{k}, \delta \mathbf{u}_{k,i}\right) \forall i \!\in \! \{0, \!\cdots\!, \!M-1\!\},
\end{equation} 
\begin{equation}\label{eq: state-dependent cost-to-go}
\tilde{q} 
\!= \!
\underbrace{
\vphantom{
\gamma_{\mathbf{u}} \delta \mathbf{u}_{k}^\top R \delta \mathbf{u}_{k}+\mathbf{u}_{k}^\top R \delta \mathbf{u}_{k}+\frac{1}{2} \mathbf{u}_{k}^\top R \mathbf{u}_{k}}
q\left(\mathbf{x}_{k}\!\right)}_{\text{\color{black}{\textit{State-dep.}}}} 
\!+ 
\underbrace{\gamma_{\mathbf{u}} \delta \mathbf{u}_{k,i}^\top R \delta \mathbf{u}_{k,i}\!+ \mathbf{u}_{k}^\top R \delta \mathbf{u}_{k,i}\!+ \frac{1}{2} \mathbf{u}_{k}^\top R \mathbf{u}_{k}}_{\text{\color{black}{\textit{$q\left(\mathbf{u}_{k}, \delta \mathbf{u}_{k} \right)$: Quadratic Control Cost}}}}.
\end{equation}

 As outlined in (\ref{eq:mppi_optimal-control}) from \cite{williams2017model}, the optimal control sequence $\left\{\mathbf{u}_{k}\right\}_{k=0}^{N-1}$ in the vanilla MPPI algorithm is iteratively updated by taking a weighted average cost from all simulated trajectories, where $\tilde{S}\left(\tau_{m}\right)$ represents the \textit{cost-to-go} of the $m^{th}$ trajectory, and $\lambda \in \mathbb{R}^{+}$ denotes the ``inverse temperature'', which regulates the selectiveness of the weighted average of the trajectories.
 After smoothing the resulting control sequence with a Savitzky-Galoy filter \cite{savitzky1964smoothing}, the first control $\mathbf{u}_{0}$ is executed in the system, with the remaining sequence utilized as a warm-start for the next optimization step. Formally, 
\begin{equation}\label{eq:mppi_optimal-control}
 \mathbf{u}_{k} \leftarrow \mathbf{u}_{k} +\frac{\sum_{m=0}^{M-1} \exp \Bigl( \frac{-1}{\lambda} \tilde{S}\left(\tau_{m}\right) \Bigr) \delta \mathbf{u}_{k, m}}{\sum_{m=0}^{M-1} \exp \Bigl( \frac{-1}{\lambda} \tilde{S}\left(\tau_{m}\right) \Bigr)}.
 \end{equation}
\subsection{Sparse Gaussian Process} 
\label{GP_prelim}
Gaussian Process (GP) is a well-established non-parametric model described by a mean function $m(\textbf{z})$ and a co-variance function $k(\textbf{z}, \textbf{z}^{\prime})$ 
(also referred to as kernel function), where $\textbf{z} \in \mathbb{R}^{n_{g}}$ is the input to the GP~\cite{GPforML}; it can be mathematically expressed as
\begin{equation}
    f(\mathbf{z}) \sim \mathcal{G P}\left(m(\mathbf{z}), k\left(\mathbf{z}, \mathbf{z}^{\prime}\right)\right).
    \label{eq_full_gp}
\end{equation} 
Let $\mathcal{D} = \left\{\left(\mathbf{z}_{i}, y_{i}\right)\right\}_{i=1}^{n}$ denote a dataset consisting of $n$ input-output pairs, where each output $y_i \in \mathbb{R}$ is assumed to be the sum of an unknown underlying function $f(\mathbf{z}_i)$ and Gaussian noise $\epsilon_i$ with a zero-mean and variance $\sigma^2$, i.e., $\epsilon_i \sim \mathcal{N}\left(0, \sigma^{2}\right)$.
In the context of GP regression, to estimate the output $y^*$  for a given new input $\textbf{z}^*$, the following GP prediction equation is employed
\begin{equation}
    \begin{aligned}
    p(y^* | \textbf{y}) &= \mathcal{N}(y^* | m_{\textbf{y}}(\textbf{z}^*), k_\textbf{y}(\textbf{z}^*,\textbf{z}^*) + \sigma^2), \\
        m_{\mathbf{y}}(\mathbf{z}) &=K_{\mathbf{z} n}\left(\sigma^{2} I+K_{n n}\right)^{-1} \mathbf{y}, \\
    k_{\mathbf{y}}\left(\mathbf{z}, \mathbf{z}^{\prime}\right) &=k\left(\mathbf{z}, \mathbf{z}^{\prime}\right)-K_{\mathbf{z} n}\left(\sigma^{2} I+K_{n n}\right)^{-1} K_{n \mathbf{z}^{\prime}},
     \end{aligned}
    \label{eq_predictive_eq_full_gp}
\end{equation} 
where $m_{\mathbf{y}}(\mathbf{z})$ and $k_{\mathbf{y}}(\textbf{z},\textbf{z}^{\prime})$ are the GP posterior mean and co-variance functions, respectively, while $K_{nn} \in \mathbb{R}^{n \times n}$ refers to the $n \times n$ co-variance matrix of the training inputs and $K_{\mathbf{z}n}\in \mathbb{R}^n$ is $n$-dimensional row vector of kernel function values between $\mathbf{z}$ and the training inputs, with $K_{n\mathbf{z}} = K_{\mathbf{z}n}^\top$. 
Achieving a more accurate GP prediction requires the optimization of hyper-parameters, such as kernel parameters $\Theta$ and noise variance $\sigma^2$, by maximizing the log marginal likelihood
\begin{equation}
    \log p(\mathbf{y})=\log \left[\mathcal{N}\left(\mathbf{y} \mid \mathbf{0}, \sigma^{2} I+K_{n n}\right)\right].
    \label{eq:eq_mlml_full_gp}
\end{equation} 

The standard GP can be computationally intensive due to its complexity of $\mathcal{O}(n^3)$, where $n$ represents the number of training instances. To mitigate this issue, various approximation methods, collectively known as Sparse Gaussian Process (SGP), have been developed as an alternative approach. Instead of using the complete training data, SGP employs a smaller set of $m_s$ training points, called \textit{inducing points} $Z_{m_s}$, resulting in a more efficient process and a lower computation complexity of $\mathcal{O}(n m_s^2)$ ~\cite{lawrence2003fast,snelson2006sparse,titsias2009variational2}. 
Our present work leverages the variational SGP method, proposed in~\cite{titsias2009variational2}, to approximate the true posterior of a GP $p(f|\mathbf{y})$ using an approximated variational posterior distribution $q(f,f_{m_s})$, where $f_{m_s}$ are the values of the underlying function $f$ at the \textit{inducing points} $Z_{m_s}$. 
This approximation is done by augmenting the true posterior with the variable $f_{m_s}$ such as $p(f,f_{m_s}|\mathbf{y}) = p(f|f_{m_s}) p(f_{m_s}|y)$. 
Then, the approximated variational distribution $q(f,f_{m_s})$ can be factorized in the same manner as the augmented true posterior, as follows
\begin{equation}
    q(f,f_{m_s}) = p(f|f_{m_s})\phi(f_{m_s}),
    \label{eq_approx_equal_true_posterior_SGP}
\end{equation}
where $\phi(f_{m_s})$ is an unconstrained variational distribution over $f
_{m_s}$ and $p(f|f_{m_s})$ is the conditional GP prior.
By minimizing the Kullback-Leibler (KL) divergence between the approximated and true posteriors, $\mathbb{K} \mathbb{L}[q(f, f_{m_s})||p(f|\mathbf{y})]$, the variational SGP obtains estimates of the inducing inputs $Z_{m_s}$ and hyperparameters ($\Theta, \sigma^2$).

\begin{figure*}[th!] 
\subfloat[Gazebo environment \label{fig_sgp_oc_a}]{%
  \includegraphics[width=0.19\textwidth,height=1.15in]{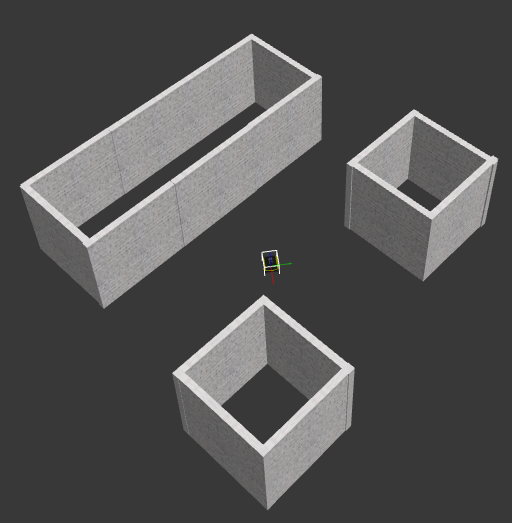} } \hfill
\subfloat[Raw pointcloud\label{fig_sgp_oc_b}]{%
  \includegraphics[width=0.19\textwidth,height=1.15in]{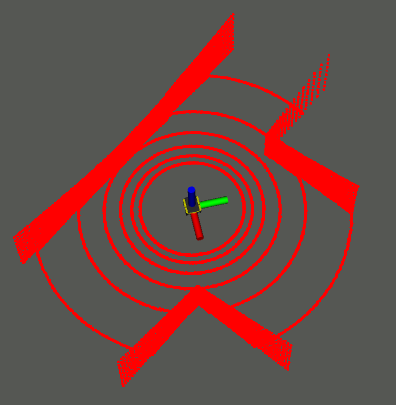} } \hfill
\subfloat[\!Original occupancy surface\label{fig_sgp_oc_c}]{%
  \includegraphics[width=0.19\textwidth,height=1.15in]{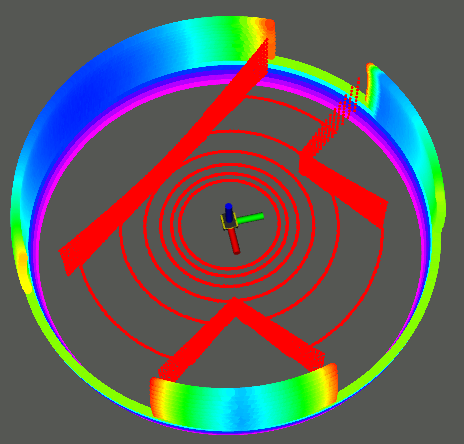}} 
  \includegraphics[width=0.01\textwidth,height=1.15in]{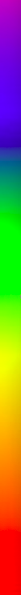} \hfill
\subfloat[SGP occupancy surface\label{fig_sgp_oc_d}]{%
  \includegraphics[width=0.19\textwidth,height=1.15in]{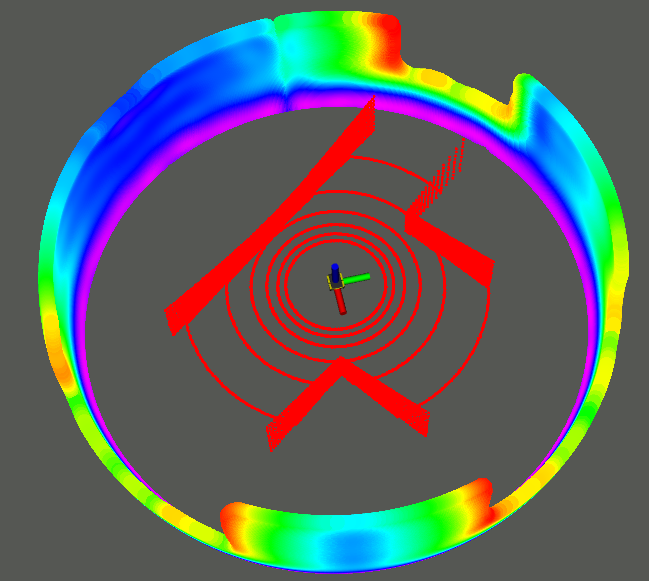}}\hfill 
\subfloat[SGP variance surface\label{fig_sgp_oc_e}]{%
  \includegraphics[width=0.19\textwidth,height=1.15in]{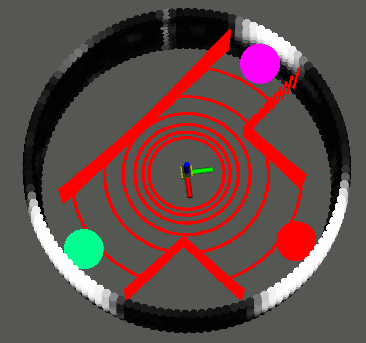}}\hfill
  \vspace*{-2pt}
  \caption{Illustrative example of the SGP occupancy model, with (a) the Jackal robot located in an unknown cluttered environment, (b) the raw pointcloud built by the onboard sensor, (c) the original occupancy surface where warmer colors indicate less occupancy, (d) the SGP occupancy surface reconstructed by our model, and (e) the SGP variance surface with white regions denoting GP frontiers and colored circles representing the navigation subgoals.
  }
  \label{fig_sgp_oc}
  \vspace*{-14pt}
\end{figure*}
\section{GP-MPPI Control Strategy}\label{GP-MPPI Control Strategy}
The goal of our present research, as outlined in \eqref{eq:2}, is to determine the optimal control sequence $\mathbf{U}=\left\{\mathbf{u}_{k}\right\}_{k=0}^{N-1}$ that enables safe and efficient navigation of the mobile robots through complex and unknown cluttered environments, while avoiding collisions with obstacles and getting trapped in local minima.
Although the MPPI control framework, as summarized in \cite{mohamed2021sampling}, has many positive attributes, it is prone to generating \textit{infeasible} control sequences or trajectories, particularly when the distribution of all sampled trajectories are concentrated within high-cost regions. 
To mitigate this issue, new sampling strategies proposed in \cite{yin2022trajectory, mohamed2022autonomous} have enabled more efficient exploration of the state-space, allowing the algorithm to find better solutions and potentially reduce the risk of trapping in local minima.
Nevertheless, for specific tasks such as the one depicted in Fig.~\ref{maze_env}, eliminating the local minima remains a potential challenge that needs to be tackled.

One solution could be incorporating MPPI with a global planner, such as the solution presented in \cite{tao2023rrt}, which utilizes the RRT algorithm to guide MPPI.
Instead, we introduce the GP-MPPI control strategy, a new \textit{online} navigation technique 
that leverages the SGP occupancy model to learn about the navigable space surrounding the robot. Specifically, we introduce the GP-subgoal recommender policy, which identifies a set of recommended subgoals and subsequently suggests the optimal subgoal that minimizes a predefined cost function to the MPPI local planner, as depicted in Fig.~\ref{fig:indoor-cluttered-enviroment} and explained in detail in Section~\ref{GP-Subgoal Recommender Policy}.
Unlike conventional methods, a distinctive aspect of the proposed control strategy is that it does not require either a global map for long-term planning or an offline training process.
\subsection{SGP Occupancy Surface Representation}\label{Occupancy Surface Representation with SGP}
Our proposed GP-subgoal recommendation policy relies on our earlier work presented in~\cite{230111251, gpfrontiers}, where we transformed pointcloud data into an \textit{occupancy surface} and modeled it using a Sparse Gaussian Process (SGP).
Within this approach, the {\em occupancy surface} takes the form of a 2D circular surface centered around the sensor origin and has a predefined radius of $r_{oc}$.
This surface serves as the projection space for all observed points, which are represented in spherical coordinates $(\theta_i, \alpha_i, r_i)$, where $(\theta_i, \alpha_i, r_i)$ correspond to the azimuth, elevation, and radius values of each observed point, respectively.
Each point $\mathbf{z}_i$ on the occupancy surface is defined by two attributes: the azimuth and elevation angles $\mathbf{z}_i= (\theta_i, \alpha_i)$, and assigned an {\em occupancy value} $f(\mathbf{z}_i)$ that is a function of the point radius $r_i$, such as $f(\mathbf{z}_i)=r_{oc}-r_i$. Afterward, the probability of occupancy {$f(\mathbf{z})$ over the occupancy surface is modeled} by an SGP occupancy model, as follows
\begin{equation}
    \begin{aligned}
    f(\mathbf{z}) &\sim \mathcal{SGP}\left(m(\mathbf{z}), k\left(\mathbf{z}, \mathbf{z}^{\prime}\right)\right),  \\
    k\left(\mathbf{z}, \mathbf{z}^{\prime}\right) &=\sigma_f^{2}\left(1+\frac{\left(\mathbf{z}-\mathbf{z}^{\prime}\right)^{2}}{2 \alpha \ell^{2}}\right)^{\!\!-\alpha}\!\!,
    \end{aligned}
    \label{eq_mean_kernel_SGP}    
\end{equation}
where $\sigma_{f}^{2}$ is the signal variance, $l$ is  the length-scale, and $\alpha$ is the relative weighting factor that manipulates large and small scale variations.
In our SGP model, the point's occupancy to radius relation is encoded as a zero-mean function, $m(\mathbf{z})=0$, where the occupancy value of the non-observed points is set to zero.
The Rational Quadratic (RQ) kernel, $k\left(\mathbf{z}, \mathbf{z}^{\prime}\right)$, is selected as the SGP kernel due to its ability to model functions that vary across different length-scale~\cite{GPforML}. This characteristic makes the RQ kernel well-suited for modeling the occupancy surface.

In Fig.~\ref{fig_sgp_oc}, we present a concrete example of the SGP occupancy model applied to our Jackal robot, which is equipped with a Velodyne VLP-16 LiDAR and located in an unknown cluttered environment, as depicted in Fig~\ref{fig_sgp_oc_a}. The figure also illustrates the raw pointcloud generated by the onboard sensor (Fig~\ref{fig_sgp_oc_b}), as well as the original occupancy surface, which represents the projection of the point clouds onto the 2D circular surface with radius $r_{oc}$, where warmer colors indicate areas of lower occupancy (Fig~\ref{fig_sgp_oc_c}). Furthermore, Fig~\ref{fig_sgp_oc_d} exhibits the SGP occupancy surface reconstructed by the SGP occupancy model, as previously expressed in~\eqref{eq_mean_kernel_SGP}.
 The precision of the SGP occupancy model is intensively evaluated in  our previous work~\cite{230111251}, where the results showed that an SGP occupancy model comprising of 400 inducing points generates a reconstructed point cloud with an average error of approximately \SI{12}{\centi\metre}.

\subsection{GP-Subgoal Recommender Policy}\label{GP-Subgoal Recommender Policy}
The primary advantage of GP and its variants, compared to other modeling techniques, is their ability to provide a measure of variance, which indicates the level of uncertainty, along with a function estimate (i.e., mean).
More precisely, in the context of the occupancy surface, the SGP occupancy model prediction, as defined in \eqref{eq_predictive_eq_full_gp}, provides both mean $\mu_{oc_i}$ and variance $\sigma_{oc_i}$ values for each point on the surface, where the mean represents the expected occupancy while the variance reflects the uncertainty associated with the predicted occupancy.
Consequently, constructing the SGP occupancy surface is accompanied by an SGP variance surface that captures the uncertainty in the occupancy estimate, as depicted in Fig.~\ref{fig_sgp_oc_e}.

Within this research, we have opened up a new avenue for effectively utilizing the SGP variance surface as a reliable indicator for distinguishing between occupied and free spaces around the robot, where regions with variances higher than a certain threshold $V_{th}$ correspond to free space, while low-variance regions indicate occupied space.
In fact, the variance surface changes across observations due to variations in the number and distribution of observed points employed in the training of the SGP model.
As a result, the variance threshold $V_{th}$ is considered to be a variable that relies on the distribution of the variance across the surface and can be calculated as $V_{th}=K_m v_m$, where $K_m \in \mathbb{R}^+$ is a tuning parameter 
and $v_m$ represents the mean of the variance distribution.
To identify free navigable spaces, we define a \textit{Gaussian Process frontier} (namely, GP frontier) as the centroid point $(\theta_{i}, \alpha_{i})$ of each high variance region. These GP frontiers $\left\{f_i\right\}_{i=1}^{\mathcal{F}}$ serve as local recommended subgoals (see colored circles in Fig.~\ref{fig_sgp_oc_e}).
Unlike the well-known frontier concept introduced in~\cite{yamauchi1998frontier}, it is worth noting that our GP frontier does not rely on a global occupancy map; instead, it is extracted from the uncertainty of the current observation.

Following the identification of the GP frontiers by the SGP model, a cost function $J_{gp}$ is utilized to determine the optimal GP frontier $f^{*}$ that guides the local planner (in our case, MPPI) towards the desired state $\mathbf{x}_f$.
Our cost function $J_{gp}$, given in \eqref{eq_cost_fun}, has been established with two distinct terms. The first term, as introduced in \cite{yan2020mapless}, calculates the distance $d_{fs}$ between a GP frontier $f_i$ and the desired state $\mathbf{x}_f$. This distance criterion is used to identify the GP frontier closest to $\mathbf{x}_f$. The second term, inspired by the direction criterion proposed in~\cite{ye2009sub}, evaluates the direction $\theta_{f_i}$ of a GP frontier with respect to the robot heading. This criterion prioritizes a GP frontier that aligns better with the robot heading.
\begin{equation}
    \begin{aligned}
        J_{gp}\left(f_{i}\right) &=  k_{dst}  d_{fs} + k_{dir} \theta_{fi}^2 ,  \\
    f^{*} &=\operatorname{arg} \min _{f_{i} \in \mathcal{F}}\left(J_{gp}\left(f_{i}\right)\right), 
    \end{aligned}
    \label{eq_cost_fun}    
\end{equation}
where $k_{dst}$, $k_{dir}$ are weighting factors. The GP frontier direction $\theta_{f_i}$ is squared to indicate the absolute direction.
Finally, the local planner receives the optimal subgoal $\textbf{g}^*$, obtained by acquiring the Cartesian coordinate of the optimal GP frontier $f^*$, which leads the robot to its desired state $\mathbf{x}_f$.

\subsection{Real-Time GP-MPPI Control Algorithm}\label{Real-Time GP-MPPI Control Algorithm}
Algorithm~\ref{alg:UMPPI-Alg.} summarizes the \textit{real-time} control cycle of the GP-MPPI algorithm, which includes two primary components: the local MPPI motion planner (described earlier in Section~\ref{Overview of MPPI Control Strategy}) and the GP-subgoal recommender (explained in Section~\ref{GP-Subgoal Recommender Policy}).
Each time-step $\Delta t$, the GP policy recommends the optimal subgoal $\textbf{g}^*$, the current state is estimated, and a $M \times N$ random control variations $\delta \mathbf{u}$ are generated (lines $2:4$). Then,
$M$ trajectories are simulated in parallel, propagated from the system dynamics defined in~\eqref{dynamics_system}, and evaluated using \eqref{eq:cost-to-go} (lines $5:13$). It is noteworthy that the minimum sampled cost trajectory, denoted as $\tilde{S}_{\min}$, among all simulated trajectories prevents numerical overflow or underflow without affecting the optimality of the algorithm \cite{mohamed2020model}.
After that, the optimal control sequence $\left\{\mathbf{u}_{k}\right\}_{k=0}^{N-1}$ is updated, smoothed with a Savitzky-Galoy filter, and the first control $\mathbf{u}_{0}$ is applied to the system (lines $14:18$), while the remaining sequence of length $N - 1$ is slid down to be utilized at next time-step (lines $19:22$).
In lines $25$ to $38$, the function known as \textit{GP-SubgoalRecommender} is described, which takes a pointcloud input (PCL) and returns the optimal subgoal $\textbf{g}^*$ for the local planner.
To optimize the hyper-parameters $\Theta$ and inducing points $Z_{m_s}$ of the SGP occupancy model, the pointcloud data is transformed into training data $\mathcal{D}$ (lines $26:29$). 
The mean occupancy $\mu_{oc}$ and variance $\sigma_{oc}$ are then estimated over the surface $Z^*$, and the GP frontiers are defined as those with $\sigma_{oc} > V_{th}$, where the centroids of these frontiers are converted to Cartesian coordinates (lines $30:34$). Finally, the cost function $J_{gp}$ in~\eqref{eq_cost_fun} is used to select the optimal subgoal $\textbf{g}^*$ (lines $35:37$).

In this study, we introduce two operating modes for the GP-MPPI algorithm: the simple mode (SM) and the recovery mode (RM). Under the simple mode, MPPI consistently leverages the optimal subgoal $\mathbf{g}^*$ suggested by the GP policy. In contrast, in the recovery mode, MPPI generates the optimal control sequence that steers the robot towards its desired state $\mathbf{x}_f$, adhering to the recommended subgoal only when the robot is at risk of encountering local minima. Such local minima occur when the robot's linear velocity is zero ($v=0$) and its current state $\mathbf{x}_k$ does not match $\mathbf{x}_f$ (i.e., $\mathbf{x}_k \neq \mathbf{x}_f$). 
Thanks to the optimal control sequence $\left\{\mathbf{u}_{k}\right\}_{k=0}^{N-1}$ obtained by MPPI, we can efficiently anticipate the occurrence of local minima by imposing a condition on the mean of the predicted linear velocities over the time-horizon $N$, expressed as follows:
\begin{equation}\label{recovery-condition}
 \mu_{\mathbf{u}} = \frac{1}{N} \sum_{i=0}^{N-1} |v_i| < \mathbf{u}_{th},    
\end{equation}
where $\mathbf{u}_{th} \in \mathbb{R}^+$ represents a control switching threshold set based on $N$.
If this condition is fulfilled, then MPPI will follow the subgoal recommended by the GP rather than navigating directly towards its desired state $\mathbf{x}_f$.
\begin{algorithm}[ht!]
\caption{Real-Time GP-MPPI Control Algorithm}
\label{alg:UMPPI-Alg.}
\hspace*{\algorithmicindent} \textbf{Given:} \\
\hspace*{1cm} $M, N$: Number of rollouts (samples) \& timesteps \\
\hspace*{1cm} $ \left(\mathbf{u}_{0}, \mathbf{u}_{1}, \ldots,
\mathbf{u}_{N-1}\right) \equiv \mathbf{U}$: Initial control sequence \\
\hspace*{1cm} $ f, \Delta t$: Dynamics \& time-step size\\
\hspace*{1cm} $\phi, q, \lambda, \nu, \Sigma_{\mathbf{u}}, Q, R$: Cost \& control hyper-parameters \\
\hspace*{1cm} SGF: Savitzky-Galoy (SG) convolutional filter \\
\hspace*{1cm} $\textit{PCL}, \mathbf{Z^*}$: Pointcloud \& 2D variance surface (Grid)\\
\hspace*{1cm} $\Theta, m_s, k_{dst}, k_{dir}, k_{m}$: GP policy hyper-parameters

\begin{algorithmic}[1] 
\While {\textit{task not completed}} 
    \State $g^* \leftarrow$ \textit{GP-SubgoalRecommender(PCL)},
    \State $\mathbf{x}_{0} \leftarrow$ \textit{StateEstimator()}, \hfill $\mathbf{x}_{0} \in \mathbb{R}^{n_x}$ 
    \State $ \delta \mathbf{u} \leftarrow$ \textit{GaussianNoiseGenerator()}, \hfill $\delta \mathbf{u} \in \mathbb{R}^{M \times N}$
    \For{$m \leftarrow 0$ \textbf{to} $M-1$ \textit{in parallel}}
        \State $\mathbf{x} \leftarrow \mathbf{x}_{0}, \quad \tilde{S}\left(\tau_{m}\right) \leftarrow  0, \hfill \tilde{S}\left(\tau_{m}\right) \in \mathbb{R}^{+}$
        \For{$k \leftarrow 0$ \textbf{to} $N-1$}
            \State $\mathbf{x}_{k+1} \leftarrow \mathbf{x}_{k}+ f\left(\mathbf{x}_{k}, \mathbf{u}_{k}+\delta \mathbf{u}_{k,m}\right) \Delta t$,
            \State $\tilde{S}\left(\tau_{m}\right) \leftarrow \tilde{S}\left(\tau_{m}\right)+\tilde{q}$,
        \EndFor
        \State $\tilde{S}\left(\tau_{m}\right) \leftarrow \tilde{S}\left(\tau_{m}\right)+ \phi\left(\mathbf{x}_{N}\right)$,  
    \EndFor
    \State $\tilde{S}_{\min} \leftarrow \min _{m}[\tilde{S}\left(\tau_{m}\right)]$, \hfill $\forall m=\{0, \dots, M-1\}$
    \For{$k \leftarrow 0$ \textbf{to} $N-1$}
        \State $\mathbf{u}_{k} \leftarrow \mathbf{u}_{k}+\frac{\sum_{m=0}^{M-1} \exp \bigl( \frac{-1}{\lambda} \left[\tilde{S}\left(\tau_{m}\right) -\tilde{S}_{\min} \right] \bigr) \delta \mathbf{u}_{k, m}}{\sum_{m=0}^{M-1} \exp \bigl(\frac{-1}{\lambda} \left[\tilde{S}\left(\tau_{m}\right) -\tilde{S}_{\min} \right]\bigr)}$,
    \EndFor 
    \State $\mathbf{u} \leftarrow \textit{SGF}(\mathbf{u})$,
    \State $\mathbf{u}_{0} \leftarrow$ \textit{SendToActuators}($\mathbf{u}$),
    \For{$k \leftarrow 1$ \textbf{to} $N-1$}
        \State $\mathbf{u}_{k-1} \leftarrow \mathbf{u}_{k}$,
    \EndFor
    \State $\mathbf{u}_{N-1} \leftarrow$ \textit{ControlSequenceInitializer}($\mathbf{u}_{N-1}$),
    \State Check for task completion
\EndWhile
\Function{\textit{GP-SubgoalRecommender}}{\textit{PCL}},
    \State ($\theta_i, \alpha_i, r_i$) $\gets$ \textit{Cartesian2Spherical(PCL($x_i, y_i, z_i$))},
    \State $ oc_i=r_{oc}-r_i$, \quad $ \mathcal{D}= \left\{\left(\mathbf{z}_{i}, oc_{i}\right)\right\}_{i=1}^{n}$,
    \State $f(\mathbf{z}) \sim \mathcal{SGP}\left(m(\mathbf{z}), k\left(\mathbf{z}, \mathbf{z}^{\prime}\right)\right)$, \quad $k \gets RQ$, 
    \State \textit{Optimize} \quad ($\Theta, Z_{m_s}$) $\gets \mathcal{D}$,
    \State $(\mu_{oc} , \sigma_{oc})\gets \mathcal{SGP}\textit{-Predict}(Z^*)$,
    \State $v_m\gets \textit{Mean}(\sigma_{oc})$, \quad $V_{th}\gets K_m v_m$,
    \State $\textit{GP-Frontiers} \gets (\sigma_{oc} > V_{th})$,
        \State $(\theta_{f_i}, \alpha_{f_i}) \gets \textit{CentroidOfGP-Frontiers}$,    
        \State $(x_{f_i}, y_{f_i}, 0) \gets$ \textit{Spherical2Cartesian}($\theta_{f_i}, \alpha_{f_i}, r_{oc}$),
        \State $d_{fs}\gets$ \textit{EuclideanDistance}($(x_{f_i}, y_{f_i}), \mathbf{x}_{f}$),
        \State $f^{*} =\operatorname{arg}\min _{f_{i} \in \mathcal{F}}\left(J_{gp}\left(f_{i}\right)\right)$,
        \State $\textbf{g}^* \gets (x_{f^*}, y_{f^*}, \theta_{f^*})$,
\State \Return $\textbf{g}^*$
\EndFunction
\end{algorithmic}
\end{algorithm}
 \vspace*{-10pt}
\section{Simulation-Based Evaluation}\label{Simulation Details and Results}
In this section, the effectiveness of our proposed control strategy is assessed and compared with both vanilla MPPI and log-MPPI control strategies in a goal-oriented autonomous ground vehicle (AGV) navigation task conducted in 2D cluttered environments of unknown nature. 
\subsection{Simulation Setup:}\label{Simulation Setup:Cluttered Environments}
In this study, we consider the kinematics model of a differential wheeled robot presented in \cite{mohamed2022autonomous}, specifically the fully autonomous ClearPath Jackal robot, where the robot's position and orientation in the world frame are given by $\mathbf{x} = [{x}, {y}, \theta]^\top \in \mathbb{R}^{3}$, and the control input $\mathbf{u} = [v,\omega]^\top \in \mathbb{R}^{2}$ denotes the robot's linear and angular velocities.
Our autonomous AGV platform is equipped with a 16-beam Velodyne LiDAR sensor utilized for two key functions: (i) constructing the SGP variance surface, and (ii) generating the local costmap.

The simulations for all proposed control schemes were conducted with the following parameters: a prediction time of \SI{6}{\second}, a control frequency of \SI{30}{\hertz} (i.e., $N=180$), sampling \num{2528} rollouts per time-step $\Delta t$, and an exploration variance $\nu$ of \num{1200}. Additionally, a control weighting matrix $R$, expressed as $\lambda \Sigma_{n}^{-\frac{1}{2}}$, is utilized. 
In the case of MPPI and GP-MPPI, the inverse temperature $\lambda$ and the control noise co-variance $\Sigma_\mathbf{u} = \Sigma_{n} = \operatorname{Diag}\left(\sigma_v^2, \sigma_w^2\right)$ are both set to \num{0.572} and $\operatorname{Diag}\left(0.023, 0.028\right)$, respectively. However, for log-MPPI, different values of \num{0.169} and $\operatorname{Diag}\left(0.017, 0.019\right)$ are used for these parameters, along with a normal distribution that has a co-variance of $\Sigma_{n} = \operatorname{Diag}\left(0.002, 0.0022\right)$ (For more details, refer to \cite{mohamed2022autonomous}).
The Savitzky-Galoy (SG) convolutional filter is utilized with a quadratic polynomial function, i.e., $n_{sg}=2$, and a window length $l_{sg}$ of $51$.
The occupancy surface was constructed with an occupancy radius $r_{oc}$ of \num{5} meters, a full azimuth range of $-180^o$ \textit{to} $180^o$, 
and elevation height of $0^o$ \textit{to} $15^o$. The SGP occupancy model was designed with \num{400} inducing points ($Z_m = 400$), where the GP frontiers were identified based on a variance threshold of $V_{th}= K_m v_m$, where $K_m$ was set to \num{0.4}.
For the distance and direction factors $K_{dst}$ and $K_{dir}$ of the cost function $J_{gp}$, we assigned weighting factors of \num{5} and \num{4}, respectively.
To enable the recovery mode of the GP-MPPI, we have set the control threshold, $\mathbf{u}_{th}$, to \SI{0.55}{[\metre\per \sec]}.
All the proposed control schemes, which are written in Python and integrated with the Robot Operating System (ROS) framework, are executed in \textit{real-time} on an NVIDIA GeForce GTX 1660 Ti laptop GPU, with the GP-subgoal recommender built on \textit{GPflow}\cite{matthews2017gpflow}.

To accomplish the 2D navigation task, we adopt a state-dependent cost function described in \eqref{eq:state-dep-cost-function}, which comprises two terms. The first term, with $Q = \operatorname{Diag}(2.5,2.5,5)$, aims to steer the robot towards its desired state, whereas the second term incorporates a Boolean variable $\mathbb{I}_{\text{crash}}$ to heavily penalizes collisions with obstacles.
\begin{equation}\label{eq:state-dep-cost-function}
q(\mathbf{x}_k)= (\mathbf{x}_k-\mathbf{x}_{f})^{\top} Q (\mathbf{x}_k-\mathbf{x}_{f}) + 10^3 \mathbb{I}_{\text{crash}}.
\end{equation}
Since the robot is operating in unknown environments, it relies on a 2D costmap to maintain a record of obstacles in its vicinity. This costmap is generated by analyzing sensor data from the environment and constructing a 2D occupancy grid, with each cell typically categorized as \textit{occupied}, \textit{free}, or \textit{unknown} \cite{costmap2016}. 
The generated occupancy grid is subsequently employed as a 2D local costmap, feeding directly into the sampling-based MPC algorithm, enabling safe and collision-free navigation.
The robot-centered \textit{2D} local costmap, which is built by the on-board Velodyne VLP-16 LiDAR sensor, has a size of $\SI{200}{\cell} \times \SI{200}{\cell}$ and a grid resolution of \SI{0.05}{\metre/\cell}.
Finally, throughout the simulations, the maximum linear velocity $v_\text{max}$ of the robot is set to \SI{1.5}{\metre/\second}.
\subsection{Simulation Scenarios and Performance Metrics:}
\begin{figure}[!ht] 
  \subfloat[Gazebo forest-like environment\label{forest_env}]{%
        \vspace*{-3pt}
        \hspace*{-8pt}\includegraphics[scale=1]{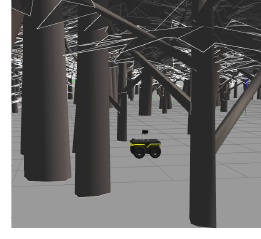}
        }
   \subfloat[Gazebo maze-like environment\label{maze_env}]{%
      \vspace*{-3pt}
      \hspace*{-9pt}\includegraphics[scale=1]{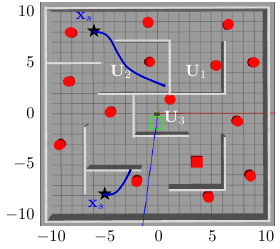}
      } 
  \caption{\small Snapshot of our Jackal robot located in (a) forest-like and (b) maze-like environments, with blue lines denoting the robot's paths generated by log-MPPI, which eventually guide it to local minima.}
  \label{Gazebo simulation environments}
  \vspace*{-5pt}
\end{figure}

The benchmark evaluation utilizes two types of Gazebo simulation environments, as depicted in Fig.~\ref{Gazebo simulation environments}. The first type, referred to as \textit{Forest \#1}, is a $\SI{50}{\metre} \times \SI{50}{\metre}$ forest-like environment characterized by tree-shaped obstacles with a density of \SI{0.2}{\trees/\square \metre}; 
The other type, named \textit{Maze \#1}, is a $\SI{20}{\metre} \times \SI{20}{\metre}$ maze-like environment with three $\text{U}$-shaped rooms (i.e., $\text{U}_1$, $\text{U}_2$, and $\text{U}_3$), as well as various other obstacles (highlighted in red in Fig.~\ref{maze_env})\footnote{To evaluate the local planner's obstacle avoidance capability, the red obstacles are intentionally made undetectable as occupied space by the GP-subgoal recommender, as occupancy elevation height is set to a higher value.}.
In the first scenario, denoted as \textit{Forest \#1}, the robot is directed to navigate from an initial pose $\mathbf{x}_s = [-5,-8,0]^\top$ to a desired pose $\mathbf{x}_f = [20,20,45]^\top$ in ([\si{\metre}], [\si{\metre}], [\si{\deg}]). 
Meanwhile, in \textit{Maze \#1}, we conducted two separate control missions to (i) evaluate the robustness of our proposed control strategy, and (ii) examine its performance under the two different operating modes, previously described in Section~\ref{Real-Time GP-MPPI Control Algorithm}.
The first mission, $\text{MU}_1$, requires the robot to navigate from $\mathbf{x}_s = [-5,-8,60]^\top$ to a desired pose $\mathbf{x}_f = [4,4,45]^\top$ located inside $\text{U}_1$; while, in the second mission, named $\text{MU}_2$, the robot starts at $\mathbf{x}_s = [-6,8,0]^\top$, crosses $\text{U}_2$, and reaches a desired pose of $\mathbf{x}_f = [8,-8,170]^\top$.
\vspace*{-15pt}

To ensure a fair and comprehensive comparison of the three control schemes, we have established a set of performance metrics, including the {\em task completion percentage} $\mathcal{T}_{\text{c}}$, the {\em average distance} traveled by the robot $d_{\text{av}}$ to reach $\mathbf{x}_{f}$ from $\mathbf{x}_{s}$, the {\em average  linear velocity} $v_{\text{av}}$ of the robot within the cluttered environment, and the {\em percentage of assistance} $\mathcal{A}_{\text{gp}}$ provided by the GP-subgoal recommender policy to MPPI when the recovery mode is utilized. The successful task completion entails the robot reaching the target position without encountering obstacles or getting trapped in local minima $\mathcal{R}_{\text{lm}}$.
\subsection{Simulation Results:}
We evaluated the effectiveness of the proposed control strategies in \textit{Forest \#1} and \textit{Maze \#1} (i.e., $\text{MU}_1$ \& $\text{MU}_2$) through 10 trials each, and the resulting performance statistics are summarized in Table~\ref{table:Comparison-maze-like-enviroment}.
The performance results demonstrate that, as expected, the proposed GP-MPPI control strategy outperforms both the vanilla MPPI and log-MPPI as the autonomous vehicle successfully accomplished all control missions (with $\mathcal{T}_{\text{c}}=100\%$) without getting stuck in local minima or colliding with obstacles (i.e., $\mathcal{R}_{\text{lm}} =0 $), despite having limited perception range and incomplete knowledge of the environment.
In contrast, in \textit{Forest \#1}, log-MPPI achieved a task completion rate $\mathcal{T}_{\text{c}}$ of 95.72\% over 10 trials, compared to 86.87\% when MPPI was utilized. Additionally, log-MPPI encountered local minima only twice, while MPPI was trapped six times.
Nevertheless, both control methods were unable to complete any of the trials in $\text{MU}_1$ and $\text{MU}_2$ due to the challenging environmental conditions (refer to the robot trajectories generated by log-MPPI in Fig.~\ref{maze_env}).
Additionally, our proposed approach in \textit{Forest \#1} provided a shorter route towards the desired state $\mathbf{x}_f$, especially when the recovery mode (RM) is activated, similar to the optimal trajectory of the baselines, with an average linear velocity $v_{\text{av}}$ of \SI{1.30}{\metre/\second}, which approaches the maximum specified velocity of $\SI{1.5}{\metre/\second}$.
\begin{table}[!ht]
\vspace*{-10pt}
\caption{
Comparison of performance statistics for proposed control strategies performing missions in \textit{Forest \#1} and \textit{Maze \#1} ($\text{MU}_1$ and $\text{MU}_2$), where the gray cells indicate better results.}
\vspace{-5pt}
\begin{center}
\small\addtolength{\tabcolsep}{-6pt} 
 \begin{tabular}{|l||c|c|c|c|}
 \hline
& \multicolumn{2}{c|}{\textit{\textbf{Baselines}}} & \multicolumn{2}{c|}{\textit{\textbf{Ours (GP-MPPI)}}}\\ 
 \cline{2-5}
 \multirow{-2}{*}{ \hspace*{-2pt}\textit{Indicator}}  & \multicolumn{1}{c|}{\textit{MPPI}} & \multicolumn{1}{c|}{\textit{log-MPPI}} & \multicolumn{1}{c|}{\textit{Mode: SM}} & \multicolumn{1}{c|}{\textit{Mode: RM}}\\
 \hline\hline
 \multicolumn{5}{|c|}{\textbf{Mission}: \textit{Forest \#1}, $v_\text{max} = \SI{1.5}{\metre/\second}$} \\
 \hline
  $\mathcal{T}_{\text{c}}$ [\%] $(\!\mathcal{R}_{\text{lm}}\!)$ & $86.87$ (6) &  $95.72$ (2) &  \cellcolor{gray!20} 100 (0) &    \cellcolor{gray!20}  100 (0) \\
 $d_{\text{av}}$ {[\si{\metre}]}&  $\cellcolor{gray!20}  38.60\pm0.04$ &  $38.62\pm0.39$ &  $ 40.19\pm0.54$ &  $39.16\pm0.35$\\
$v_{\text{av}}$ [\si{\metre/\second}] &  $1.29\pm 0.01$  &  $1.18\pm0.24$    &  $1.29\pm0.01$  & \cellcolor{gray!20} $1.30\pm0.01$ \\
$\mathcal{A}_{\text{gp}} [\%]$     & $-$&   $-$ &   $-$ &       $11.04\pm3.58$\\
 \hline
 \multicolumn{5}{|c|}{\textbf{Mission}: $\text{MU}_1$, $v_\text{max} = \SI{1.5}{\metre/\second}$} \\
 \hline
 $\mathcal{T}_{\text{c}} [\%]$   $(\!\mathcal{R}_{\text{lm}}\!)$    & 10.93 (10) &  10.96 (10) & \cellcolor{gray!20} 100 (0) &    \cellcolor{gray!20}  100 (0) \\
 $d_{\text{av}}$ {[\si{\metre}]}  & $3.58\pm0.03$&  $3.59\pm0.06$& $34.48\pm1.85$ & \cellcolor{gray!20}  $32.74\pm0.41$\\
$\mathcal{A}_{\text{gp}} [\%]$& $-$ & $-$ &  $-$ &   $46.82\pm5.33$\\
\hline
 \multicolumn{5}{|c|}{\textbf{Mission}: $\text{MU}_2$, $v_\text{max} = \SI{1.5}{\metre/\second}$} \\
 \hline
 $\mathcal{T}_{\text{c}} [\%]$  $(\!\mathcal{R}_{\text{lm}}\!)$    & 22.66 (10) & 22.76  (10) &  \cellcolor{gray!20} 100 (0) &   \cellcolor{gray!20}    100 (0) \\
 $d_{\text{av}}$ {[\si{\metre}]} &  $8.82\pm0.08$ &  $8.86\pm0.09$ &  \cellcolor{gray!20} $38.92\pm1.11$ &  $41.74\pm1.36$\\
$\mathcal{A}_{\text{gp}} [\%]$     & $-$ &   $-$ &   $-$ &       $45.88\pm4.89$ \\
 \hline
\end{tabular}
\end{center}
\label{table:Comparison-maze-like-enviroment}
\vspace*{-8pt}
\end{table}

\begin{figure}[!ht]
  \subfloat[\textit{Maze \#1}: $\text{MU}_1$\label{maze_path1}]{%
      \vspace*{-3pt}
      \hspace*{-6pt}\includegraphics[scale=1]{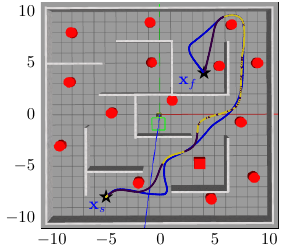}
      }
   \subfloat[\textit{Maze \#1}: $\text{MU}_2$\label{maze_path2}]{%
      \vspace*{-3pt}
      \hspace*{-10pt}\includegraphics[scale=1]{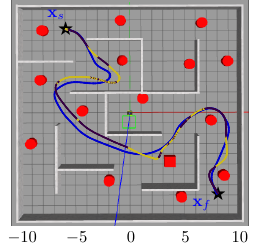}
      } 
  \caption{
  Behavior of GP-MPPI in a $\SI{20}{\metre} \times \SI{20}{\metre}$ maze-like environment under the two operating modes, where the blue trajectory indicates the simple mode; in the recovery mode, the yellow color indicates the trajectory's segments where GP assisted MPPI to avoid local minima.
  }
  \label{maze_path1_and_2}
  \vspace*{-22pt}
\end{figure}

\begin{figure}[!ht] 
  \subfloat[Linear velocity $v$ \label{maze_path1_velocity}]{%
      \vspace*{-3pt}
      \includegraphics[scale=1]{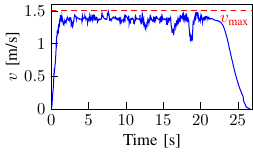}
      }
   \subfloat[Mean of linear velocities $\mu_{\mathbf{u}}$ \label{maze_path1_mean_velocity}]{%
      \vspace*{-3pt}
      \hspace*{-5pt}\includegraphics[scale=1]{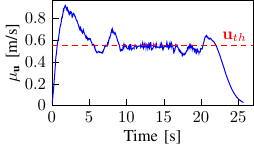}
      } 
  \caption{Robot velocity profile and the corresponding mean of predicted linear velocities for GP-MPPI with RM on the $\text{MU}_1$ mission.}
  \label{maze_path1_velocit_profile}
  \vspace*{-5pt}
\end{figure}

Concerning the two modes of GP-MPPI, it is observed that activating the recovery mode (RM) during \textit{Forest \#1} and $\text{MU}_1$ missions improves the average distance traveled $d_{\text{av}}$ by the robot. For instance, in $\text{MU}_1$, $d_{\text{av}}$ was approximately \SI{32.74}{\metre} with RM, whereas with the simple mode (SM), which consistently relies on the subgoal recommended by GP, $d_{\text{av}}$ was roughly \SI{34.48}{\metre}. On the other hand, during the $\text{MU}_2$ mission, the RM produced a slightly longer robot trajectory than the SM since operating our proposed GP-MPPI in the RM strikes a balance between the state-dependent cost function that directs the robot to follow a direct route towards the desired state and the optimal subgoal recommended by the GP policy that forces the robot to avoid the dead-ends associated with rooms $\text{U}_2$ and $\text{U}_3$ on its way to $\mathbf{x}_f$, as illustrated in Fig.~\ref{maze_path2}.
We can also see that, due to the presence of U-shaped rooms in \textit{Maze \#1}, the GP provides more assistance, represented by $\mathcal{A}_{\text{gp}}$, than in \textit{Forest \#1}.
In Fig.~\ref{maze_path1_and_2}, we illustrate through an example from the conducted trials the robot trajectories generated by GP-MPPI under the two operating modes in \textit{Maze \#1}. We can clearly observe that our proposed control strategy successfully achieves collision-free navigation in both modes, without getting stuck in local minima.
As an example, Fig.~\ref{maze_path1_velocity} displays the velocity profile of the robot during the $\text{MU}_1$ mission shown in Fig.~\ref{maze_path1}, while using GP-MPPI with RM, along with its corresponding mean of the predicted linear velocities $\mu_\mathbf{u}$ over the given time-horizon $N$ (see Fig.~\ref{maze_path1_mean_velocity}). The mean values that fall below the switching threshold $\mathbf{u}_{th}$, set at \SI{0.55}{[\metre \per\sec]}, denote the intervals where the RM is active, and are visually emphasized in yellow in Fig.~\ref{maze_path1}.
\vspace*{-5pt}
 \vspace*{-5pt}
\section{Real-World Demonstration}\label{Real-World Demonstration}
In this section, we experimentally demonstrate the applicability of our proposed control strategy in achieving a safe 2D grid-based collision-free navigation in a complex and unknown indoor cluttered environment.
\subsubsection{Experimental Setup and Validation Environment:}\label{Experimental Setup:real-world Environment}
To conduct our experimental validation, we used the simulation setup previously outlined in Section~\ref{Simulation Setup:Cluttered Environments}, except for (i) setting the maximum speed $v_\text{max}$ to \SI{1.0}{\metre/\second} to avoid the robot localization error associated with using the RealSense camera as a source of localization, (ii) setting the occupancy radius $r_{oc}$ to \SI{3.0}{\metre}, 
and (iii) decreasing the size of the 2D grid map to $\SI{120}{\cell} \times \SI{120}{\cell}$.
\setlength{\intextsep}{0pt} 
\begin{wrapfigure}{r}{0.25\textwidth}
\begin{center}
\includegraphics[scale=0.12]{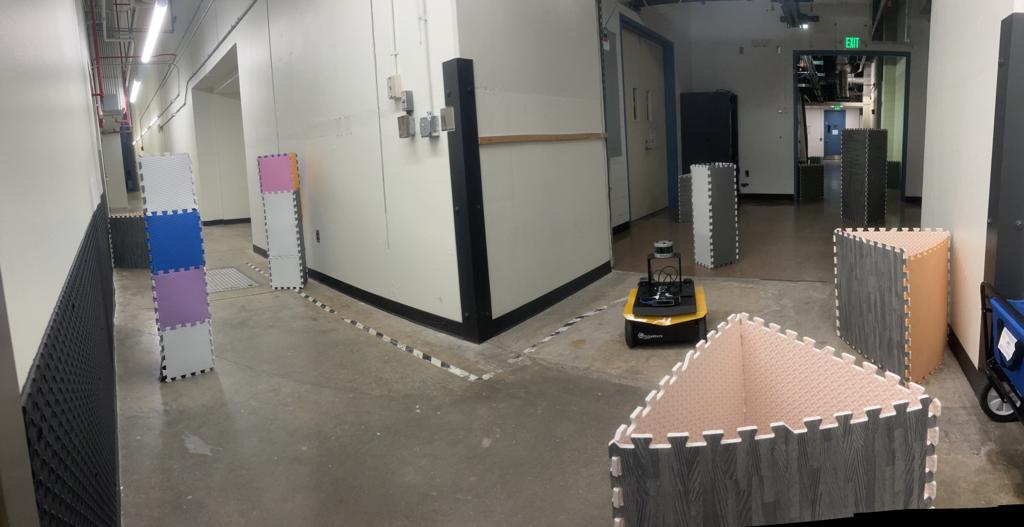}
\caption{Panoramic photo of our L-shaped indoor environment.} 
\label{fig:L-shaped indoor enviroment}
\end{center}
\end{wrapfigure}
\setlength{\intextsep}{0pt}
We also decreased the recovery mode switching threshold $\mathbf{u}_{th}$ to \SI{0.3}{\metre/\second} to be compatible with the updated $v_\text{max}$.
Additionally, to ensure \textit{real-time} execution of the GP-subgoal recommender policy, we decrease the resolution of the SGP variance surface to one-third of its original value along the azimuth axis while keeping the original resolution along the elevation axis.
We employed an L-shaped indoor corridor environment measuring $\SI{9}{\metre} \times \SI{14}{\metre}$ for experimental validation. The environment has a varying width between \SI{1.8}{\metre} and \SI{2.8}{\metre} and contains randomly placed boxes-like obstacles, as depicted in Fig.~\ref{fig:L-shaped indoor enviroment}. 
The assigned control mission of the robot is to navigate from $\mathbf{x}_s = [0,0,0]^\top$~and arrive at $\mathbf{x}_f = [7.5,13,90]^\top$.

\subsubsection{Experimental Results:}\label{Experimental results:real-world Environment}  
The performance statistics of our proposed GP-MPPI control scheme, gathered from four trials conducted in our indoor environment, are summarized in Table~\ref{table:real-world-results} for the two operating modes.
From all trials, we can conclude that both operating modes provide collision-free navigation in the cluttered environment with an average linear velocity of \SI{0.80}{\metre \per \sec}, without the risk of being trapped in local minima (as $\mathcal{R}_{\text{lm}} = 0$) while moving towards its desired state. This ensures the safety and consistent feasibility of the receding-horizon planning.
In contrast, it is observed that the vanilla MPPI and log-MPPI consistently failed to complete any of the trials due to being trapped in the first edge of the L-shaped environment. 
However, MPPI managed to avoid such traps with the aid of the GP-subgoal recommender policy in the recovery mode (RM), which provides an average assistance percentage $\mathcal{A}_{\text{gp}}$ of roughly 31.36\%.
More details about the simulation and experimental results, including the behavior of the baselines, are provided in the supplementary video: \url{https://youtu.be/et9t8X1wHKI}.
 \vspace*{5pt}
\begin{table}[!ht]
\caption{
Performance statistics of the two modes of GP-MPPI.
}
\vspace{-6pt}
\begin{center}
\small\addtolength{\tabcolsep}{-4pt} 
 \begin{tabular}{|c||c|c|c|c|c|}
 \hline
Mode & $\mathcal{T}_{\text{c}}$ [\%] $(\!\mathcal{R}_{\text{lm}}\!)$  & $d_{\text{av}}$ [\si{\metre}] & $v_{\text{av}}$ [\si{\metre \per \sec}] & $\mathcal{A}_{\text{gp}} [\%]$\\
\hline
\textit{SM} & 100 (0)  & \cellcolor{gray!20} $20.06\pm0.21$    & \cellcolor{gray!20} $0.80\pm 0.012$  & $-$  \\
\textit{RM} & 100 (0)  & $20.18\pm0.30$    & $0.76\pm 0.066$  & $31.36\pm11.72$ \\
 \hline
\end{tabular}
\end{center}
\label{table:real-world-results}
\vspace*{-4pt}
\end{table}
\section{Conclusion}\label{sec:conclusion}
 In this work, we proposed the GP-MPPI control strategy, which comprises two primary components: the GP-subgoal recommender policy and the local planner, the MPPI. First, the GP-subgoal recommender utilized the learning capacity of SGP to create a reliable SGP variance surface, which served as an indicator for differentiating between occupied and free spaces around the robot. Consequently, a set of suggested subgoals was identified, and the optimal subgoal that minimizes a predefined cost function was recommended to the local MPPI planner. Based on the recommended subgoal, MPPI computes the optimal control input that enables the robot to navigate towards the goal efficiently and safely while accounting for its dynamics and avoiding collisions.
By conducting a combination of simulated and real-world experiments, we have shown that our proposed control strategy is superior to the vanilla MPPI and log-MPPI methods in achieving efficient and safe navigation in unknown and complex environments, thereby avoiding the risk of getting stuck in local minima.
\bibliographystyle{IEEEtran}
\bibliography{references}            

\end{document}